\documentclass[10pt,twocolumn,letterpaper]{article}

\usepackage{cvpr}
\usepackage{times}
\usepackage{epsfig}
\usepackage{graphicx}
\usepackage{amsmath}
\usepackage{amssymb}
\usepackage{calrsfs}
\usepackage{subcaption}
\graphicspath{{./figures/}}
\DeclareMathAlphabet{\pazocal}{OMS}{zplm}{m}{n}

\DeclareMathOperator*{\argmin}{arg\,min}
\usepackage[ruled,vlined]{algorithm2e}

\newcommand{\Lb}{\pazocal{L}}
\usepackage{siunitx}
\usepackage[breaklinks=true,bookmarks=false]{hyperref}

\hypersetup{
  colorlinks   = true, %Colours links instead of ugly boxes
  urlcolor     = magenta, %Colour for external hyperlinks
  linkcolor    = red, %Colour of internal links
  citecolor   = red %Colour of citations
}

\cvprfinalcopy % *** Uncomment this line for the final submission

\makeatletter
\newcommand{\printfnsymbol}[1]{%
  \textsuperscript{\@fnsymbol{#1}}%
}
\makeatother

\def\cvprPaperID{****} % *** Enter the CVPR Paper ID here

\begin{document}

%%%%%%%%% TITLE
\title{Preventing Clean Label Poisoning using Gaussian Mixture Loss}

\author{Muhammad Yaseen\thanks{denotes equal contribution.}\hspace{10mm} Muneeb Aadil\printfnsymbol{1} \hspace{10mm} Maria Sargsyan\printfnsymbol{1}\\
Max Planck Institute for Informatics, Saarbr\"ucken, Germany\\
{\tt\small \{myaseen, maadil, msargsyan\}@mpi-inf.mpg.de}
}

\maketitle 
%\thispagestyle{empty}

%%%%%%%%% ABSTRACT
\begin{abstract}
Since 2014 when Szegedy et al. \cite{first} showed that carefully designed perturbations of the input can lead Deep Neural Networks (DNNs) to wrongly classify its label, there has been an ongoing research to make DNNs more robust to such malicious perturbations. In this work, we consider a poisoning attack called Clean Labeling poisoning attack (CLPA) \cite{poisonfrogs}. The goal of CLPA is to inject seemingly benign instances which can drastically change decision boundary of the DNN due to which subsequent queries at test time can be mis-classified. We argue that a strong defense against CLPA can be embedded into the model during the training by imposing features of the network to follow a Large Margin Gaussian Mixture distribution in the penultimate layer. By having such a prior knowledge, we can systematically evaluate how unusual the example is, given the label it is claiming to be. We demonstrate our builtin defense via experiments on MNIST and CIFAR datasets. We train two models on each dataset: one trained via softmax, another via LGM \cite{RethinkingFD}. We show that using LGM \cite{RethinkingFD} can substantially reduce the effectiveness of CLPA while having no additional overhead of data sanitization. The code to reproduce our results is available online.

%used to show how vulnerable the model can become to Clean-Label Poisoning Attack \cite{poisonfrogs} when using ordinary SoftMax loss. The second model is augmented with the defense and is trained so that the features on the last layer follow a mixture-of-Gaussian with L-GM loss \cite{RethinkingFD}. For each of the datasets, we show that the model trained with L-GM loss is more robust to the Clean-Label Poisoning attack \cite{poisonfrogs}.   
\end{abstract}

%%%%%%%%% BODY TEXT
\section{Introduction}

\begin{figure}[t]
\begin{center}
  \includegraphics[width=0.90\linewidth]{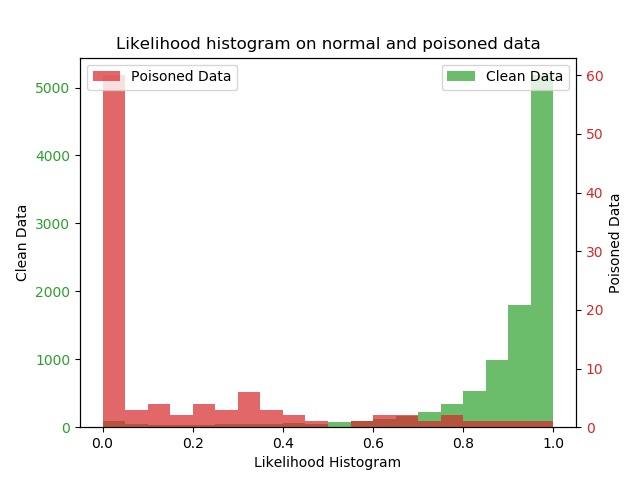}
\end{center}
\caption{Distribution of likelihood values for cleaned (MNIST Test Set) and poisoned data: Poisoned and cleaned instances have mostly low and high likelihood, respectively. This suggests we can differentiate confidently b/w cleaned and poisoned instances via likelihood thresholding.}
\label{fig:demo}
\end{figure}

With the ubiquity of Deep Neural Networks (DNNs), the issues concerning their security are becoming more and more relevant. There is thus an increasing interest in the research community to create Neural Networks which achieve state-of-the-art results, while also being secure, private, and robust. It is well known that DNNs are vulnerable to adversarial perturbations \cite{first} and an adversary might corrupt the input imperceptibly but maliciously which would change the classification result.

One such class of adversarial attacks is clean-label poisoning attacks (CLPA). In CLPA, an attacker constructs a poison training instances which looks like one class to human, but like another class to DNN. The aftermath of poisoning is that the attacker can then, during test time, can query the DNN with the malignant class which could be mis-classified as benign. Thus, effectively, a malignant instance can surpass the security mechanism. We try to tackle such an attack exploiting the intuition that specially constructed poisoning instances are far away from the class distribution in the feature space of the class they're claiming to be. That is why, we impose Gaussian Mixture distribution on the features. Experimentally, we show that it is relatively difficult to generate adversarial examples for our model. To our knowledge, this is the first model which embeds the CLPA defense into the network itself without requiring additional overhead of data sanitization. 

%-------------------------------------------------------------------------
\section{Background}
In this section, we explain the background relevant to our model. First, we formally define CLPA \cite{poisonfrogs}. Secondly, we review LGM \cite{RethinkingFD} and explain how likelihood of the features can be computed. 

\subsection{Clean-Label Poisoning attacks} \label{sec:clpa-review}

In \cite{poisonfrogs} authors introduced the notion of \textit{Clean Labelling Poisoning Attack} (CLPA) which we describe below. \newline

Let Alice be an adversary, Bob be a potential victim. Suppose that Charlie trains a huge network $F(x)$ on a gigantic cloud dataset $D_{C}$ and uploads the weights online. Further assume that Bob wishes to finetune the model $F(x)$ for some task for which the clean finetuning dataset $D_{f} = (X_{i}, Y_{i})_{i=1}^{{n}_{f}}$ is available online. However, Alice constructs a poisoned dataset $D_{p} = (X_{i}, Y_{i})_{i=1}^{{n}_{p}}$ and uploads it amid the $D_{f}$ to construct total finetuning dataset $D_{t} = D_{f} \cup D_{p}$. Notice that Bob is unaware of the poisoned instances, since he will download the available data online (which also potentially includes $D_{p}$). Next, Bob will train his model $f'(x)$ on $D_{t}$ which can potentially alter the otherwise reasonable decision boundary into vulnerable one, allowing subsequent target (usually malignant) class to be misclassified into base (usually benign) class (see figure \ref{fig:clpa-demo} in appendix \ref{app:clpa-demo}). \newline

In \cite{poisonfrogs}, the authors show how easy it is to make a clean-label targeted attack to the class of models trained by transfer learning techniques just with a single crafted examples. More specifically, to create the poisoned examples, authors optimize the following objective:

\begin{equation}\label{poisonfrog eq}
    \bold{p} = \argmin_x \| f(\bold{x})-f(\bold{t}) \|^2_2 + \beta \| \bold{x}-\bold{b} \|_2^2
\end{equation}

Where $t$ and $b$ are target and base image respectively. $f(x)$ represents the activations of penultimate layer, and $\beta$ is a trade-off parameter. As such, the \textit{image} of constructed poison is similar to base instance, while its \textit{features} resemble that of target instance. 

\subsection{Gaussian Mixture Loss}

\begin{figure}[t]
\centering
\begin{subfigure}[b]{0.475\linewidth}
        \includegraphics[width=\linewidth]{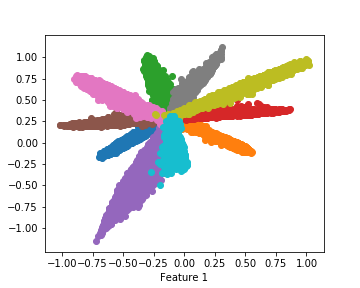}
        \caption{Softmax Loss}
        \label{fig:softmax-loss-clean-dist}
    \end{subfigure}
    \begin{subfigure}[b]{0.475\linewidth}
        \includegraphics[width=\linewidth]{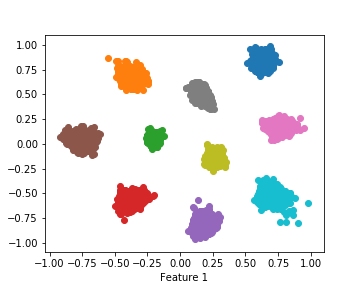}
        \caption{LGM Loss ($\lambda = 1$)}
        \label{fig:lgm-loss-clean-dist}
    \end{subfigure}
\caption{Feature Distribution of MNIST Training Set: Features are far apart for LGM Loss than for Softmax Loss. Different classes are color coded.}
\label{fig:feat-clean-dist}
\end{figure}

\begin{figure*}
\centering
\begin{subfigure}[b]{.425\textwidth}
\centering
\includegraphics[width=.21\textwidth]{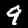}\quad
\includegraphics[width=.21\textwidth]{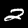}\quad
\includegraphics[width=.21\textwidth]{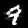}\quad
\includegraphics[width=.21\textwidth]{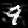}\quad
\end{subfigure}\quad \hspace{2cm}
\begin{subfigure}[b]{.425\textwidth}
\centering
\includegraphics[width=.21\textwidth]{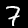}\quad
\includegraphics[width=.21\textwidth]{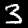}\quad
\includegraphics[width=.21\textwidth]{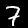}\quad
\includegraphics[width=.21\textwidth]{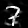}\quad
\end{subfigure} \vspace{.5cm}

\begin{subfigure}[b]{.425\textwidth}
\centering
\includegraphics[width=.21\textwidth]{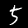}\quad
\includegraphics[width=.21\textwidth]{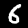}\quad
\includegraphics[width=.21\textwidth]{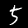}\quad
\includegraphics[width=.21\textwidth]{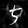}\quad
\end{subfigure}\quad \hspace{2cm}
\begin{subfigure}[b]{.425\textwidth}
\centering
\includegraphics[width=.21\textwidth]{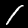}\quad
\includegraphics[width=.21\textwidth]{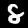}\quad
\includegraphics[width=.21\textwidth]{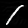}\quad
\includegraphics[width=.21\textwidth]{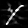}\quad
\end{subfigure}

\caption{Comparison of Constructed Poisons on MNIST Test Set for Softmax and LGM. Each set of 4 pictures contains (from left to right): $b$ (base), $t$ (target), $p_{CE}$ (poison for softmax), $p_{LGM}$ (poison for LGM). For each set, notice that $P_{CE}$ is imperceptible to human, unlike $P_{LGM}$ which has noticeable artifacts. This suggests that for $F_{LGM}(x)$, constructing imperceptible poisons is relatively more challenging.}
\label{fig:poisons}
\end{figure*}

In \cite{RethinkingFD} authors introduced a new approach to make feature distributions more structured by incorporating some prior assumptions about features in the loss function. They assume the features of penultimate layer to be realizations from a mixture of $K$ Gaussians corresponding to $K$ classes. During training, this Gaussian structure is enforced by incorporating a loss term which measures the deviation from distribution and penalizes proportionally. The model learns the means of $K$ Gaussians as parameters which is encouraged to have large inter-distribution distance ($\alpha$). For simplicity, we used only isotropic mixture of Gaussians in this project i.e. the co-variance matrix is set to identity. 

The modelling assumptions are as follows: Features $\boldsymbol{x}$ of a particular class are assumed to have the density shown in Eq \ref{feature_dist_eq}. The class conditional density i.e. distribution of features give the label is then given by Eq \ref{feature_dist_cond}. The posterior probability of class given feature is thus obtained via Bayes rule as shown in Eq \ref{posterior}. This enables us to get the likelihood of an example belonging to a class given its features.

\begin{equation}\label{feature_dist_eq}
    p(\boldsymbol{x}) = \sum_{k=1}^{K} \mathcal{N}(x; \mu_k,\,\Sigma_k)p(k)
\end{equation}

\begin{equation}\label{feature_dist_cond}
    p(\boldsymbol{x_i}|\boldsymbol{z_i}) = \mathcal{N}(x_i; \mu_{z_i},\,\Sigma_{z_i})
\end{equation}

\begin{equation}
    \label{posterior}
    p(\boldsymbol{z_i}|\boldsymbol{x_i}) = 
    \frac{\mathcal{N}(x_i; \mu_{z_i},\,\Sigma_{z_i}) p(z_i)}{   \sum_{k=1}^{K} \mathcal{N}(x; \mu_k,\,\Sigma_k)p(k)}
\end{equation}
Under the assumptions described above, the large-margin Gaussian mixture loss is given in Eq \ref{loss_main}. It consists of two components: (1) $\Lb_{cls}$ (softmax loss) and (2) $\Lb_{lkd}$ (deviation from Gaussian distribution) and $\lambda$ is a trade-off parameter.

\begin{equation}\label{loss_main}
   \Lb_{GM} = \Lb_{cls} + \lambda \Lb_{lkd}
\end{equation}

For further details, we refer the reader to \cite{RethinkingFD}.

% \begin{equation}\label{loss_lkd}
%   \Lb_{cls} = - \frac{1}{N} \sum_{i=1}^{N} \log \frac{\mathcal{N}(x_i; \mu_{z_i},\,\Sigma_{z_i}) p(z_i)}{   \sum_{k=1}^{K} \mathcal{N}(x; \mu_k,\,\Sigma_k)p(k)}
% \end{equation}

% and,
% \begin{equation}\label{loss_cls}
%     \Lb_{lkd} = - \sum_{i=1}^{N} \log \mathcal{N}(x_i; \mu_{z_i},\,\Sigma_{z_i})
% \end{equation}

\section{Proposed Method}

As explained in section \ref{sec:clpa-review}, the attacker generates a poison instance $(x_{p}, y_{p})$ such that the features $f(x_{p})$ of different classes in a softmax pre-trained model get close-by. However, since there is no way to query the likelihood $p(f(x_{p}) | y_{p})$, we cannot systematically know how ``unusual" the example $x_{p}$ is for the class $y_{p}$ that the poisoned instance is claiming to be. \newline
To this end, we use LGM loss \cite{RethinkingFD} to get likelihood $p(f(x_{p}) | y_{p})$ of an example belonging to the class $y_{p}$ it is claiming to be. The intuition is that poisoned instances $(x_{p}, y_{p})$ are far away from their claimed class in the feature space. Thus, poisoned examples will have low likelihood using which we can remove such suspicious instances before fine-tuning the model. The complete proposed procedure is highlighted in the threat model in algorithm \ref{algorithm:proposed-method}.

\begin{algorithm}
\SetAlgoLined
 \begin{enumerate}
     \item \textbf{$F_{LGM}(x)$ is pre-trained on $D_{C}$ via LGM loss}
     \item Alice generates poisoned dataset $D_{p}$
     \item $D_{p}$ is mixed along $D_{f}$ to create $D_{t} = D_{f} \cup D_{p}$
     \item Bob downloads $D_{t}$ and $F_{LGM}(x)$
     \item \textbf{Bob constructs filtered clean dataset $D_{w} = (X_{i}, Y_{i})_{i=1} \forall i,\ldots, n_{t}$ $s.t.$ $p(F_{LGM}(X_{i}) | Y_{i}) > T$}
     \item Bob can now fine-tune his model $f'(x)$ on $D_{w}$
 \end{enumerate}
 \caption{Our Contributions (Bold) in Attack Model}
 \label{algorithm:proposed-method}
\end{algorithm}

\section{Experiments and Results}

In this section, we describe our experimental details and present the results of our proposed methodology. The code to reproduce the following experiments is available online\footnote{\url{https://github.com/muneebaadil/likelihoods-for-poison}}.

\subsection{Datasets and Simulation Strategy}

\begin{figure*}
\centering
\begin{subfigure}[b]{0.475\textwidth}
        \includegraphics[width=\linewidth]{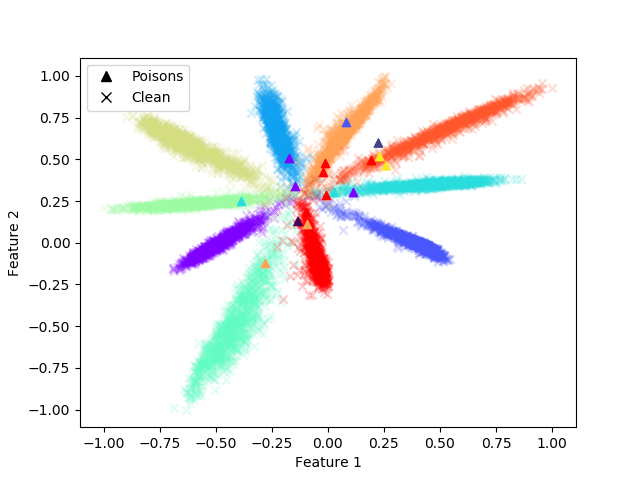}
        \caption{Softmax Loss}
        \label{fig:softmax-poisoned-dist}
    \end{subfigure}
    \begin{subfigure}[b]{0.475\textwidth}
        \includegraphics[width=\linewidth]{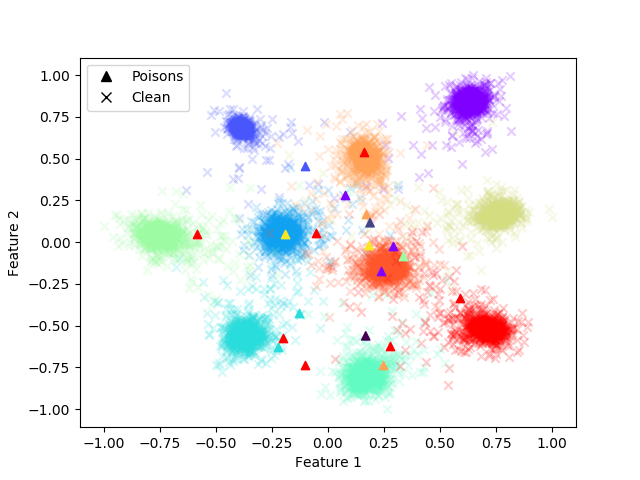}
        \caption{LGM Loss ($\lambda = 0.1$)}
        \label{fig:lgm-poisoned-dist}
    \end{subfigure}
\caption{Feature Distribution of MNIST Test Set and Poisoned Examples: Clean and poisoned instances are color coded by their ground truth class and base class respectively. Note that only 20 random poisons are shown for clarity.}
\label{fig:feat-poisoned-dist}
\end{figure*}

We use two standard datasets: MNIST \cite{lecun2010mnist}, and CIFAR10 \cite{Krizhevsky09learningmultiple}. While the original poisoning paper \cite{Shafahi2018PoisonFT} used ImageNet \cite{ILSVRC15} dataset, we skipped it because of computational constraints. \newline
Please note that in the following experiments, we treat training sets as $D_{C}$, while test sets as $D_{f}$. As such, the proposed methodology corresponds to training base networks (which can be thought of as pretrained networks $F(x)$) on $D_{C}$, while creating poisons $D_{p}$ on test sets (since in real life, attackers poison on the $D_{f}$).

\subsection{Training Base Models}
To check if training a model with LGM loss \cite{RethinkingFD} serves as a good prevention mechanism against clean labelling poisoning attacks, we train two identical CNNs; one with standard cross entropy loss ($F_{CE}(x)$) and another with LGM ($F_{LGM}(x)$). The intuition of doing so is such that if our hypothesis is correct, we should see a difficulty in generating poisons for $F_{LGM}(x)$ as opposed to $F_{CE}(x)$. These two models now can be thought of as pre-trained models $F(x)$ trained on cloud dataset $D_{C}$. Due to brevity, we do not describe the network architecture here; however, it is presented in Appendix \ref{app:architecture} for the interested readers. \newline

We train both networks until convergence; the feature distribution for both loss functions is shown in figure \ref{fig:feat-clean-dist}. Notice that features are far apart for different classes in LGM loss unlike standard cross entropy loss. Thus, loosely speaking, it should be relatively difficult than softmax to change the features of base class to resemble target class \textit{while maintaining similarity to base class in the image space}.

\subsection{Generating Poisons}

Once the base models $F_{CE}(x)$ and $F_{LGM}(x)$ are trained, we implemented poisoning algorithm according to \cite{poisonfrogs} to construct poisons for both base models to evaluate if and how much is there a difference in poisoning examples, when $F_{LGM}(x)$ is employed. \newline

As the threat model assumes that the adversary can only inject 10\% of the data into $D_{f}$, we constructed 100 poisoning instances $D_{p}$ to inject inside $D_{f}$ (which is, in our case, test-sets of MNIST/CIFAR10)\footnote{Although, 10\% of our test-sets is 1000, we only constructed 100 because of computational constraints.}. For each poisoning instance, target image $t$ and base image $b$ was chosen randomly. And similarly as in the original paper \cite{poisonfrogs}, we set $maxIters=1000$ to construct a poison instance. Lastly, we cross-validated $\beta$ parameter in the algorithm and empirically found $\beta = \num{8e-3}$ to be the best performing one.\newline

Figure \ref{fig:poisons} compares visual examples of constructed poisons for $F_{LGM}(x)$ and $F_{CE}(x)$ on MNIST.\footnote{Due to limited space, we put CIFAR10 results on Appendix \ref{app:cifar-poisons}}. Notice that for each set of base $b$ and target $t$, $p_{CE}$ is much less noticeable of an adversarial example than $p_{LGM}$, thereby suggesting $F_{LGM}$ to be more robust against poisoning. We argue this is because $F_{LGM}(x)$ features are far apart for different classes, which makes changing a feature representation \textit{without significant changes to image} challenging. Furthermore, figure \ref{fig:feat-poisoned-dist} shows the constructed poisoned instances in feature space.

\subsection{Using Likelihood to Filter out Poisons}

One direct benefit of using a structured feature representation such as Gaussian is that we can evaluate the posterior likelihood of features given the label $p(f(x)|y)$. Thus, we can ask the model that under learned representation, how likely this feature will be encountered in a class. This property of LGM can be leveraged to prevent CLPA by simply using the likelihood as an inherent trust score of an (image,label) pair and we can discard any impostor example where $p(F_{LGM}(x) | y_{claimed}) < T$.\newline

\begin{figure}[h]
\begin{center}
  %\includegraphics[width=0.9\linewidth]{roc.jpg}
  \includegraphics[width=0.9\linewidth]{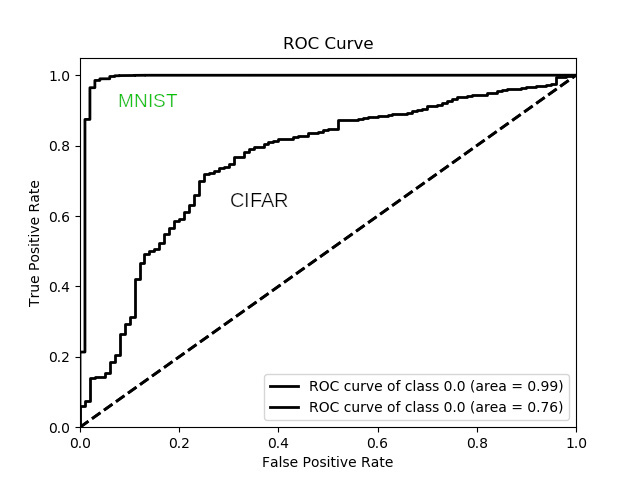}
\end{center}
\caption{ROC Curve for different likelihood thresholds to filter out poisons.}
\label{fig:roc}
\end{figure}

Figure \ref{fig:demo} shows that poisoned and normal inputs are well separated in the likelihood space and thus can be easily distinguished. This is further confirmed by the ROC plot in Figure \ref{fig:roc} plotted over different threshold levels.    

\section{Conclusion}

In this work, we showed that structured feature distributions such as mixture of Gaussians substantially restrict the effectiveness of CLPA by making it more challenging to construct clean poisons. It also additionally provides the ability to query feature likelihood which again helps in filtering the potentially poisonous examples. We demonstrate our techniques on two datasets i.e. MNIST and CIFAR-10. The constructed poisons on both datasets under LGM loss are visibly very perturbed and would fail to pass as clean labels. Additionally, we also show that even if the poison is created, network is successfully able to detect it by thresholding the feature likelihood thus preventing the attack.

\begin{figure*}
\centering

\begin{subfigure}[b]{.425\textwidth}
\centering
\includegraphics[width=.30\textwidth]{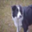}\quad
\includegraphics[width=.30\textwidth]{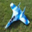}\quad
\includegraphics[width=.30\textwidth]{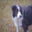}\quad
\end{subfigure}\quad \hspace{2cm}
\begin{subfigure}[b]{.425\textwidth}
\centering
\includegraphics[width=.30\textwidth]{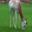}\quad
\includegraphics[width=.30\textwidth]{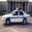}\quad
\includegraphics[width=.30\textwidth]{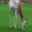}\quad
\end{subfigure} \vspace{.5cm}

\begin{subfigure}[b]{.425\textwidth}
\centering
\includegraphics[width=.30\textwidth]{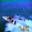}\quad
\includegraphics[width=.30\textwidth]{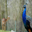}\quad
\includegraphics[width=.30\textwidth]{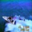}\quad
\end{subfigure}\quad \hspace{2cm}
\begin{subfigure}[b]{.425\textwidth}
\centering
\includegraphics[width=.30\textwidth]{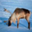}\quad
\includegraphics[width=.30\textwidth]{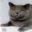}\quad
\includegraphics[width=.30\textwidth]{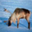}\quad
\end{subfigure}

\caption{Constructed Poisons on CIFAR Test Set for LGM. Each set of 3 pictures contains (from left to right): $b$ (base), $t$ (target), $p_{LGM}$ (poison for LGM).}
\label{fig:poisons-cifar10}
\end{figure*}

{\small
\bibliographystyle{ieee_fullname}
\bibliography{egbib}
}
\newpage
\appendix
\section{CLPA Demonstration} \label{app:clpa-demo}
\begin{figure}[h]
\begin{center}
  \includegraphics[width=0.8\linewidth]{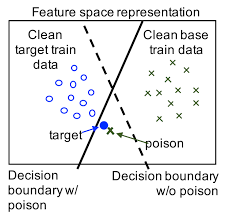}
\end{center}
\caption{Demonstration of CLPA: A crafted poison instance changes the otherwise decent decision boundary, which mis-classifies a target into base. Figure taken from \cite{poisonfrogs}}
\label{fig:clpa-demo}
\end{figure}

\section{Constructed Poisons for CIFAR10} \label{app:cifar-poisons}
\begin{figure}[h]
\begin{center}
  %\includegraphics[width=0.9\linewidth]{roc.jpg}
  \includegraphics[width=0.9\linewidth]{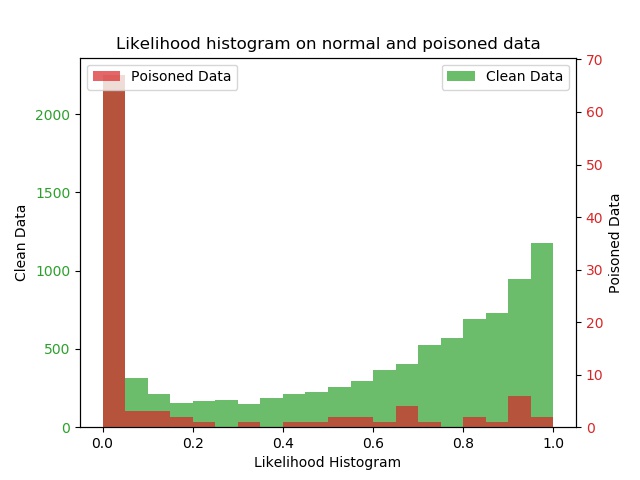}
\end{center}
\caption{Likelihood histogram for CIFAR10. Our CIFAR10 model wasn't convergent and we couldn't explore it further because of time limit.}
\label{fig:likelihood-cifar}
\end{figure}

Figure \ref{fig:poisons-cifar10} shows constructed poisons for CIFAR-10 dataset for $F_{LGM}(x)$; figure \ref{fig:likelihood-cifar} displays the likelihood statistics of clean and poisoned data.

\section{Neural Network Architecture} \label{app:architecture}
\subsection{MNIST}
% \usepackage{graphicx}

\begin{table}[]
\centering
\resizebox{\textwidth}{!}{%
\begin{tabular}{lllll}
---------------------------------------------------------------- &                      &          &  &  \\
\textbf{MNIST Architecture for LGM }                                                     & \textbf{Output Shape}         & \textbf{Param} \# &  &  \\
================================================================ &                      &          &  &  \\
Conv2d-1                                                         & {[}-1, 32, 28, 28{]} & 832      &  &  \\
PReLU-2                                                          & {[}-1, 32, 28, 28{]} & 1        &  &  \\
Conv2d-3                                                         & {[}-1, 32, 28, 28{]} & 25,632   &  &  \\
PReLU-4                                                          & {[}-1, 32, 28, 28{]} & 1        &  &  \\
MaxPool2d-5                                                      & {[}-1, 32, 14, 14{]} & 0        &  &  \\
Conv2d-6                                                         & {[}-1, 64, 14, 14{]} & 51,264   &  &  \\
PReLU-7                                                          & {[}-1, 64, 14, 14{]} & 1        &  &  \\
Conv2d-8                                                         & {[}-1, 64, 14, 14{]} & 102,464  &  &  \\
PReLU-9                                                          & {[}-1, 64, 14, 14{]} & 1        &  &  \\
MaxPool2d-10                                                     & {[}-1, 64, 7, 7{]}   & 0        &  &  \\
Conv2d-11                                                        & {[}-1, 128, 7, 7{]}  & 204,928  &  &  \\
PReLU-12                                                         & {[}-1, 128, 7, 7{]}  & 1        &  &  \\
Conv2d-13                                                        & {[}-1, 128, 7, 7{]}  & 409,728  &  &  \\
PReLU-14                                                         & {[}-1, 128, 7, 7{]}  & 1        &  &  \\
MaxPool2d-15                                                     & {[}-1, 128, 3, 3{]}  & 0        &  &  \\
Flatten-16                                                       & {[}-1, 1152{]}       & 0        &  &  \\
PReLU-17                                                         & {[}-1, 1152{]}       & 1        &  &  \\
Linear-18                                                        & {[}-1, 2{]}          & 2,306    &  &  \\
Linear-19                                                        & {[}-1, 10{]}         & 30       &  &  

\\
LGM                                                        & {[}-1, 10{]}         & 0       &  &  \\
\\
SoftMax                                                        & {[}-1, 10{]}         & 0       &  &  \\
================================================================ &                      &          &  & 
\end{tabular}%
}
\end{table}

\newpage
\subsection{CIFAR10}
% \usepackage{graphicx}
% \usepackage{graphicx}
% \usepackage{graphicx}
We used VGG19 Architecture for training CIFAR10.

\end{document}